\documentclass{article}

\usepackage[final, nonatbib]{neurips_2020}

\usepackage[utf8]{inputenc} 
\usepackage[T1]{fontenc}    
\usepackage{hyperref}       
\usepackage{url}            
\usepackage{booktabs}       
\usepackage{amsfonts}       
\usepackage{nicefrac}       
\usepackage{microtype}      
\usepackage{graphicx}
\usepackage{subcaption}

\usepackage{dsfont}
\usepackage{mathtools}
\usepackage{amsmath,amssymb}
\DeclareMathOperator{\E}{\mathbb{E}}
\DeclareMathOperator{\Var}{\mathbb{V}}
\let\P\relax
\DeclareMathOperator{\P}{\mathbb{P}}
\DeclareMathOperator{\defeq}{\dot{=}}
\DeclareMathOperator{\Cov}{\mathbb{C}}
\DeclareMathOperator{\ind}{\mathds{1}}

\usepackage{amsthm}
\newtheorem{theorem}{Theorem}

\newtheorem{assumption}{Assumption}

\usepackage[linesnumbered,ruled]{algorithm2e}

\newcommand\numberthis{\addtocounter{equation}{1}\tag{\theequation}}

\usepackage[american]{babel}
\usepackage[backend=biber, style=apa, natbib=true, sortcites=true, sorting=nyt]{biblatex}
\DeclareLanguageMapping{american}{american-apa}
\title{Hindsight Network Credit Assignment}

\author{
Kenny Young\\
Department of Computing Science\\
University of Alberta\\
Edmonton, AB, Canada\\
\texttt{kjyoung@ualberta.ca} \\
}

\begin{document}

\maketitle

\begin{abstract}
We present Hindsight Network Credit Assignment (HNCA), a novel learning method for stochastic neural networks, which works by assigning credit to each neuron's stochastic output based on how it influences the output of its immediate children in the network. We prove that HNCA provides unbiased gradient estimates while reducing variance compared to the REINFORCE estimator. We also experimentally demonstrate the advantage of HNCA over REINFORCE in a contextual bandit version of MNIST. The computational complexity of HNCA is similar to that of backpropagation. We believe that HNCA can help stimulate new ways of thinking about credit assignment in stochastic compute graphs.
\end{abstract}

The idea of using discrete stochastic neurons within neural networks is appealing for a number of reasons, including representing complex multimodal distributions, modeling discrete choices within a compute graph, providing regularization, and enabling nontrivial exploration. However, training such neurons efficiently presents challenges, as backpropagation is not directly applicable. A number of techniques have been proposed for producing either biased, or unbiased estimates of gradients for stochastic neurons. \citet{bengio2013estimating} proposes an unbiased REINFORCE~\citep{williams1992simple} style estimator, as well as a biased but low variance estimator by treating a threshold function as constant during backpropagation. \citet{tang2013learning} propose an EM procedure which maximizes variational lower bound on the loss. \citet{maddison2016concrete} and \citet{jang2016categorical} each propose a biased estimator based on a continuous relaxation of discrete outputs. \cite{tucker2017rebar} use such a continuous relaxation to derive a control variate for a REINFORCE style estimator, resulting in a variance reduced \textit{unbiased} gradient estimator.

We introduce a novel, unbiased, and computationally efficient estimator for the gradients of stochastic neurons which reduces variance by assigning credit to each neuron based on how much it impacts the outputs of its immediate children. Our technique is inspired by the recently proposed Hindsight Credit Assignment~\citep{harutyunyan2019hindsight} approach to reinforcement learning, hence we call it Hindsight Network Credit Assignment (HNCA). Aside from the immediate application to stochastic neural networks, we believe this line of thinking can help pave the way for new ways of thinking about credit assignment in stochastic compute graphs (see the work of \citet{weber2019credit} and \citet{schulman2015gradient} for some current techniques).

\textbf{Problem setting.}
We consider the problem of training a network of stochastic neurons. The network consists of a directed acyclic graph where each node is either an input node, hidden node, or output node. Each hidden node and output node correspond to a stochastic neuron that generates output according to a parameterized stochastic policy conditioned on it's incoming edges. Let $\Phi$ be a random variable corresponding to the output of a particular neuron. We define $B(\Phi)$ to be the set of random variables corresponding to outputs of the parent nodes of $\Phi$ (i.e. nodes with incoming edges to the neuron $\Phi$), we will similarly use $C(\Phi)$ to denote the children of $\Phi$. We assume each neuron's output takes a discrete set of possible values. Let $\pi_\Phi(\phi|b)$ be the policy of the neuron $\Phi$ which is defined to be equal to the probability that $\Phi=\phi$ conditioned on the neurons inputs $B(\phi)=b$. That is, $\pi_\Phi(\phi|b)\defeq\P(\Phi=\phi|B(\Phi)=b)$~\footnote{All expectations and probabilities are taken with respect to all random variables in the network, input and reward function.}. $\pi_\Phi(\phi|b)$ is a parameterized function with a set of learnable parameters $\theta_\Phi$. For simplicity, assume the network has a single output node $\hat{\Phi}$.

We focus on a contextual bandit setting, where the network selects an action conditioned on the input with the aim of maximizing an unknown reward function. The approach can generalize straightforwardly to other settings, such as supervised learning. Define the reward function $R(\Phi_0,\hat{\Phi},\epsilon)$, where $\epsilon$ is unobserved i.i.d. noise and $\Phi_0$ is a random variable corresponding to the network input. We will also use $R$ to represent the random variable corresponding to the output of the reward function. At each timestep the network receives an i.i.d. input  $\Phi_0$. The goal is to tune the network parameters to make $\E[R]$ as high as possible. Towards this, we are interested in constructing an unbiased estimator of the gradient $\frac{\partial\E[R]}{\partial\theta_\Phi}$ for the parameters of each unit, such that we can update the parameters according to the estimator to improve the reward in expectation. 

Directly computing the gradient of the output probability with respect to the parameters for a given input, as we do in backpropagation, is intractable for most stochastic networks. Computing the output probability $\P(\hat{\Phi}|\Phi_0)$ (or the associated gradient) would require marginalizing over all possible output values of each node. Instead, we can define a local REINFORCE estimator~\footnote{For Bernoulli neurons, this is equivalent to the unbiased estimator proposed by~\citet{bengio2013estimating}} as $\hat{G}^{RE}_{\Phi}=\frac{\partial\log(\pi_\Phi(\Phi|B(\Phi)))}{\partial\theta_\Phi}R$. $\hat{G}^{RE}_{\Phi}$ is an unbiased estimator of the gradient $\frac{\partial\E[R]}{\partial\theta_\Phi}$ (see Appendix~\ref{local_REIN_unbiased}). However, $\hat{G}^{RE}_{\Phi}$ tends to have significant variance.



\textbf{Hindsight Network Credit Assignment.}
We now come to the main contribution of this report, introducing HNCA: a variance reduced policy gradient estimator for learning in stochastic neural networks. HNCA works by exploiting the causal structure of the stochastic compute graph to assign credit to each node's output based on how it impacts the output of its immediate children.

To aid us in deriving HNCA, define the action value for a particular neuron as follows:
\begin{equation}
    Q_\Phi(\phi,b)=\E[R|B(\Phi)=b,\Phi=\phi].
\end{equation}
HNCA involves constructing a stochastic estimator of $Q_\Phi(\phi,b)$. That is, a random variable $\hat{Q}_\Phi(\phi)$ such that: 
\begin{equation}\label{action_value_estimator}
\E[\hat{Q}_\Phi(\phi)|B(\Phi)=b]=Q_\Phi(\phi,b).
\end{equation}
Given any such estimator, we can construct a policy gradient estimator
\begin{equation}\label{PG_estimator}
\hat{G}_{\Phi}=\sum_{\phi}\frac{\partial\pi_\Phi(\phi|B(\Phi))}{\partial\theta_\Phi}\hat{Q}_\Phi(\phi).
\end{equation}
Any such estimator is an unbiased estimator of the gradient in the sense that $\E[\hat{G}_{\Phi}]=\frac{\partial\E[R]}{\partial\theta_\Phi}$ (see Appendix~\ref{action_value_estiamtor_unbiased}). Choosing $\hat{Q}_\Phi(\phi)=\hat{Q}^{RE}_\Phi(\phi)\defeq\frac{\ind(\Phi=\phi)}{\pi_\Phi(\phi|B(\Phi))}R$ recovers the local REINFORCE estimator $\hat{G}^{RE}_{\Phi}$. Before deriving HNCA, we introduce an additional assumption:
\begin{assumption}\label{parent_of_child}
Parents of children are not descendants. More precisely, let $C^{+}(\Phi)$ be the set of descendants of $\Phi$; for every node in the network we assume that:
\begin{equation*}
    B(C(\Phi))\cap C^{+}(\Phi)=\emptyset.
\end{equation*}
\end{assumption}

This assumption guarantees that the \textit{parents of the children} of $\Phi$ provide no additional information relevant to predicting $\Phi$ given the parents of $\Phi$. This holds for typical feedforward networks as well as more general architectures. We leave open the question of whether it can be relaxed. The figures in Appendix~\ref{parent_of_child_examples} show a simple examples where Assumption~\ref{parent_of_child} holds and another where it is violated.

With this assumption in place, for any non-output node ($\Phi\neq\hat{\Phi}$), we can rewrite $Q_\Phi(\phi,b)$ as follows:
{\allowdisplaybreaks
\begin{align*}
    Q_\Phi(\phi,b)&\stackrel{(a)}{=}\E\left[\frac{\ind(\Phi=\phi)}{\pi_\Phi(\phi|B(\Phi))}R\middle|B(\Phi)=b\right]\\
    &\stackrel{(b)}{=}\E\left[\E\left[\frac{\ind(\Phi=\phi)}{\pi_\Phi(\phi|B(\Phi))}R\middle|C(\Phi),B(\Phi),B(C(\Phi))\setminus\Phi,R\right] \middle|B(\Phi)=b\right]\\
    &=\E\left[\frac{\E\left[\ind(\Phi=\phi)\middle|C(\Phi),B(\Phi),B(C(\Phi))\setminus\Phi,R\right]}{\pi_\Phi(\phi|B(\Phi))}R\middle|B(\Phi)=b\right]\\
    &\stackrel{(c)}{=}\E\left[\frac{\E\left[\ind(\Phi=\phi)\middle|C(\Phi),B(\Phi),B(C(\Phi))\setminus\Phi\right]}{\pi_\Phi(\phi|B(\Phi))}R\middle|B(\phi)=b\right]\\
    &=\E\left[\frac{\P\left(\Phi=\phi\middle|C(\Phi),B(\Phi),B(C(\Phi))\setminus\Phi\right)}{\pi_\Phi(\phi|B(\Phi))}R\middle|B(\Phi)=b\right]\\
    &\stackrel{(d)}{=}\E\left[\frac{\P\left(\Phi=\phi\middle|C(\Phi),B(\Phi),B(C(\Phi))\setminus\Phi\right)}{\P(\Phi=\phi|B(C(\Phi))\setminus\Phi,B(\Phi))}R\middle|B(\Phi)=b\right]\\
    &\stackrel{(e)}{=}\E\left[\frac{\P\left(C(\Phi)\middle|B(\Phi),B(C(\Phi))\setminus\Phi,\Phi=\phi\right)}{\P(C(\Phi)|B(C(\Phi))\setminus\Phi,B(\Phi))}R\middle|B(\Phi)=b\right]\\
    &\stackrel{(f)}{=}\E\left[\frac{\P\left(C(\Phi)\middle|B(C(\Phi))\setminus\Phi,\Phi=\phi\right)}{\P(C(\Phi)|B(C(\Phi))\setminus\Phi,B(\Phi))}R\middle|B(\Phi)=b\right],\\\numberthis\label{HNCA_derivation}
\end{align*}}
where $(a)$ uses $\hat{Q}^{RE}$; $(b)$ uses the law of total expectation; $(c)$ follows from the fact that the conditioning variables besides R form a Markov blanket~\citep{pearl2014probabilistic} for $\phi$, hence we can drop R knowing it provides no additional information; $(d)$ follows from assumption~\ref{parent_of_child}; $(e)$ applies Baye's rule; and $(f)$ follows from the fact that $B(C(\Phi))$ separates $C(\Phi)$ from $B(\Phi)$. The final expression gives rise to what we call the HNCA action-value estimator:
\begin{equation}\label{HNCA_estimator}
    \hat{Q}^{HNCA}_\Phi(\phi)=\frac{\P\left(C(\Phi)\middle|B(C(\Phi))\setminus\Phi,\Phi=\phi\right)}{\P(C(\Phi)|B(C(\Phi))\setminus\Phi,B(\Phi))}R.
\end{equation}

Effectively, HNCA assigns credit to a particular action choice $\phi$ based on the relative likelihood of it's children's actions had $\phi$ been chosen, independent of the actual value of $\Phi$. This provides a variance reduction, because an action will only get credit relative to other choices if it makes a significant difference further downstream. Note that Equation~\ref{HNCA_derivation} applies only to nodes for which $C(\Phi)\neq\emptyset$ and thus excludes the output node $\hat{\Phi}$. For $\hat{\Phi}$ we will simply use the reinforce estimator $\hat{Q}^{RE}_\Phi(\phi)$.

As stated in the following theorem, the HNCA action-value estimator always has variance lower than or equal to the local REINFORCE estimator. Let $\Var$ represent variance.
\begin{theorem}\label{reduced_variance}
$\Var(\hat{Q}^{HNCA}_{\Phi}(\Phi)|B(\phi)=b)\leq \Var(\hat{Q}^{RE}_{\Phi}(\Phi)|B(\phi)=b).$
\end{theorem}
Theorem~\ref{reduced_variance} follows from the law of total variance by the proof available in Appendix~\ref{HNCA_action_value_low_var}. In Appendix~\ref{HNCA_gradient_low_var}, we further analyze how this affects the variance of $\hat{G}^{HNCA}_{\Phi}\defeq\sum_{\phi}\frac{\partial\pi_\Phi(\phi|B(\Phi))}{\partial\theta_\Phi}\hat{Q}^{HNCA}_\Phi(\phi)$; a key result is that, for Bernoulli neurons, which we use in our experiments, we also have $\Var(\hat{G}^{HNCA}_{\Phi}(\Phi)|B(\phi)=b)\leq \Var(\hat{G}^{RE}_{\Phi}(\Phi)|B(\phi)=b)$.

\textbf{A Computationally Efficient Implementation.}
We provide a computationally efficient pseudo-code implementation of HNCA for a network consisting of Bernoulli hidden neurons with a soft-max output neuron. The implementation is similar to backprop in that each node receives information from its parents in the forward pass and passes back information to its children in a backward pass to compute a gradient estimate. Like backprop, the required compute is proportional to the number of edges in the graph. See Appendix~\ref{implementation_details} for pseudo-code and further details on this implementation.

\begin{figure*}
\centering
\begin{subfigure}{0.3\textwidth}
\includegraphics[width=\textwidth]{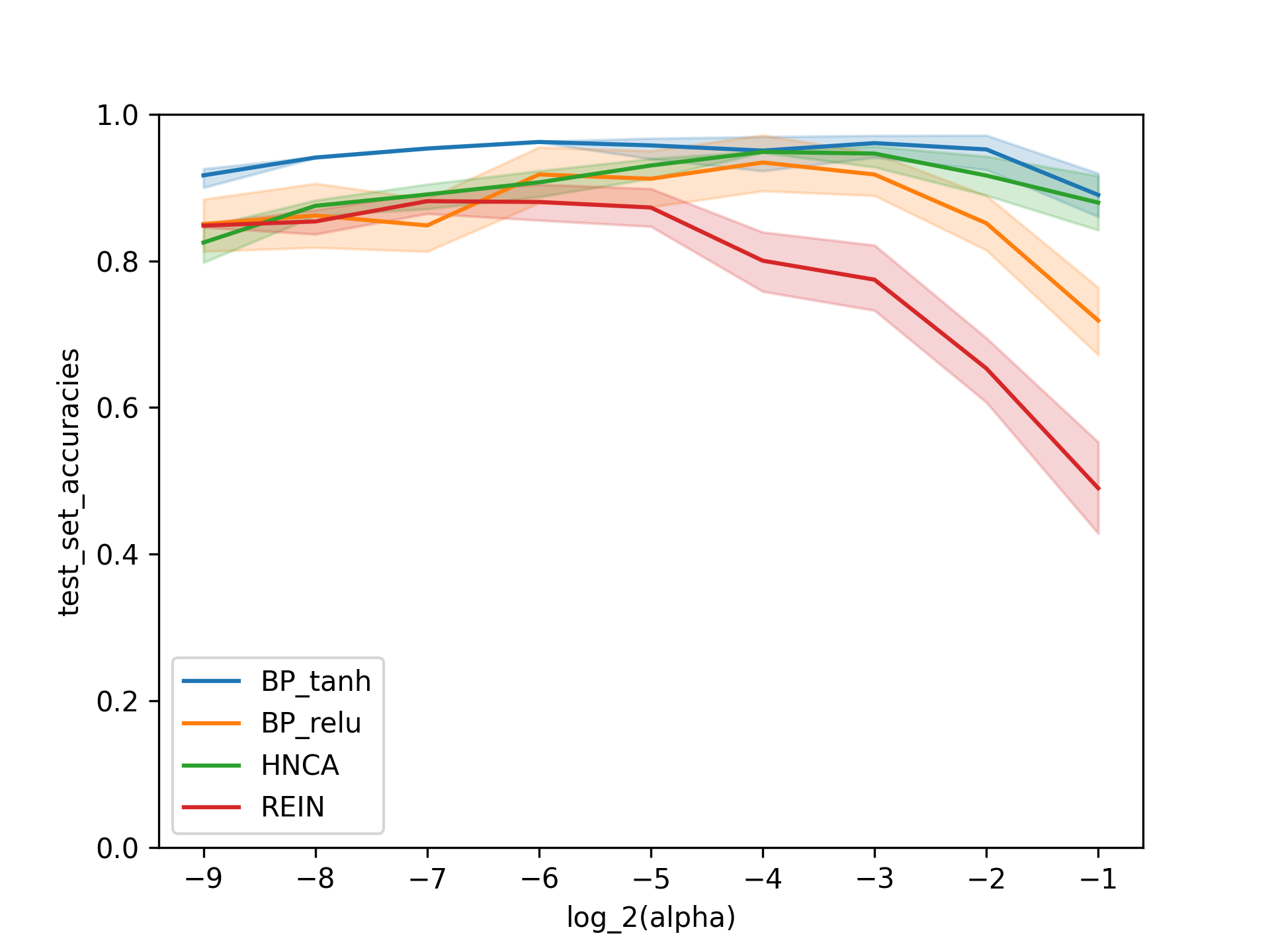}
\caption{Learning-rate sensitivity curves for 1 hidden layer.}
\label{SCE_oscillatory}
\end{subfigure}
\hfill
\begin{subfigure}{0.3\textwidth}
\includegraphics[width=\textwidth]{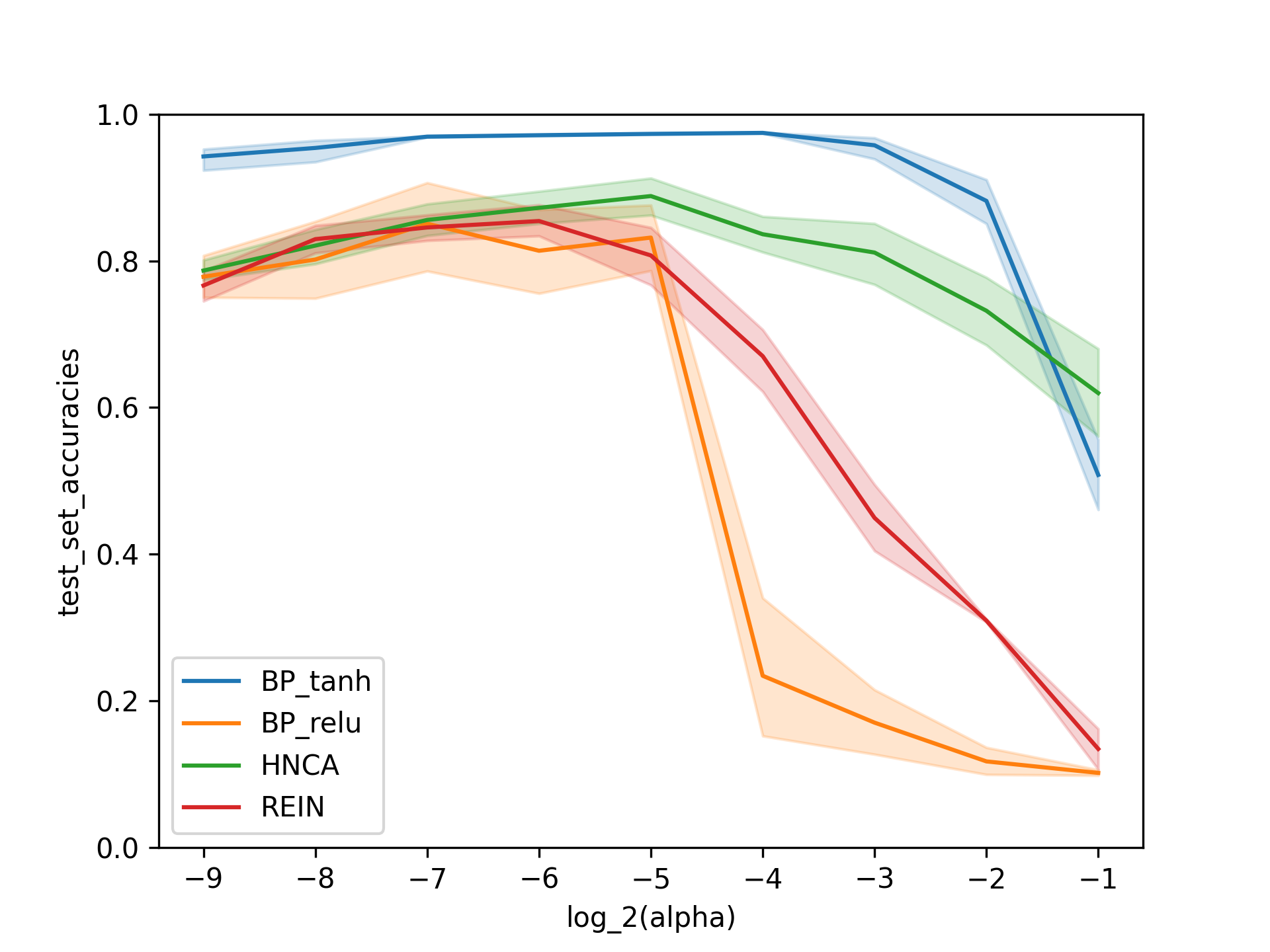}
\caption{Learning-rate sensitivity curves for 2 hidden layers.}
\label{SCE_oscillatory}
\end{subfigure}
\hfill
\begin{subfigure}{0.3\textwidth}
\includegraphics[width=\textwidth]{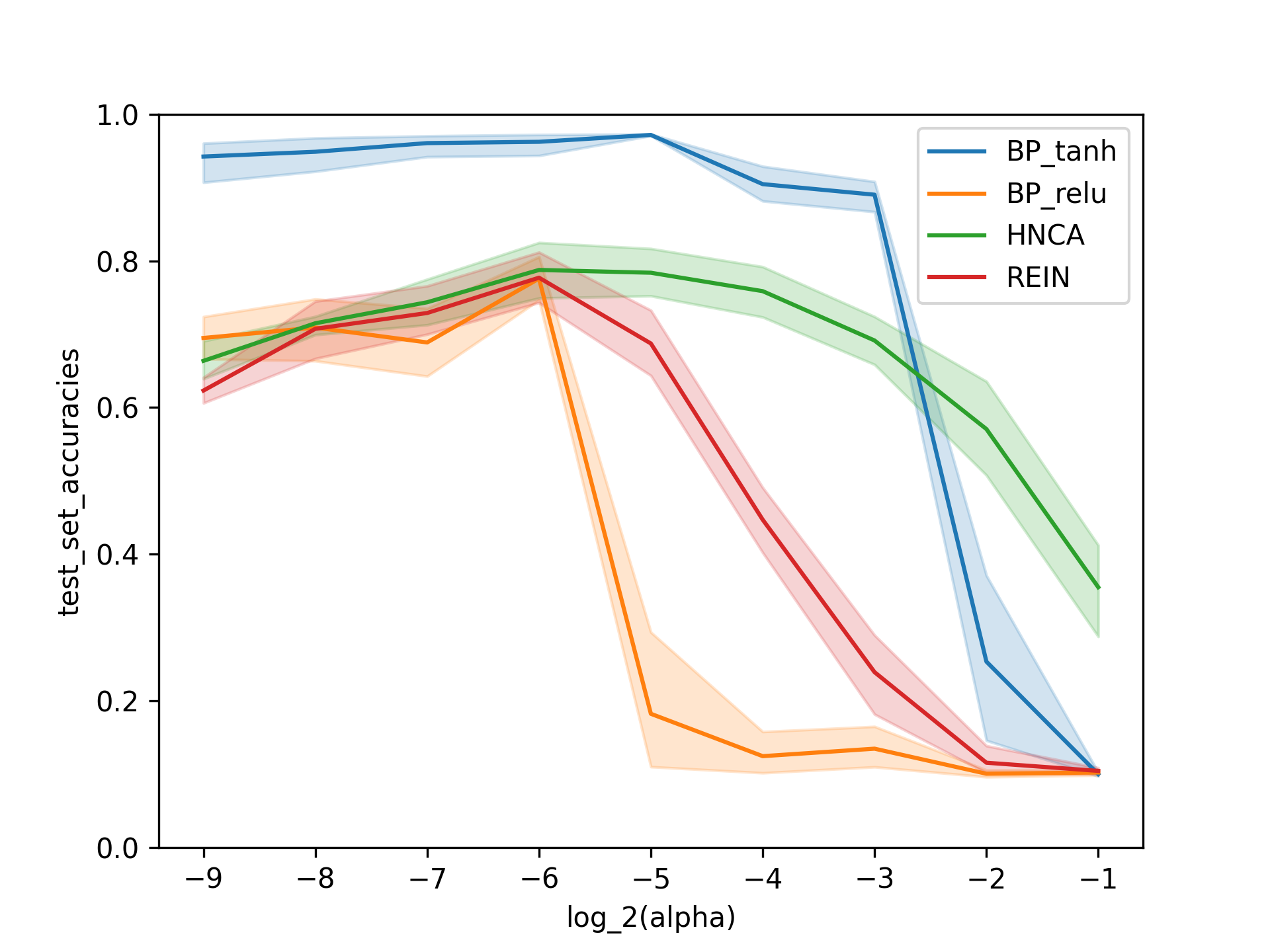}
\caption{Learning-rate sensitivity curves for 3 hidden layers.}
\end{subfigure}
\linebreak
\begin{subfigure}{0.3\textwidth}
\includegraphics[width=\textwidth]{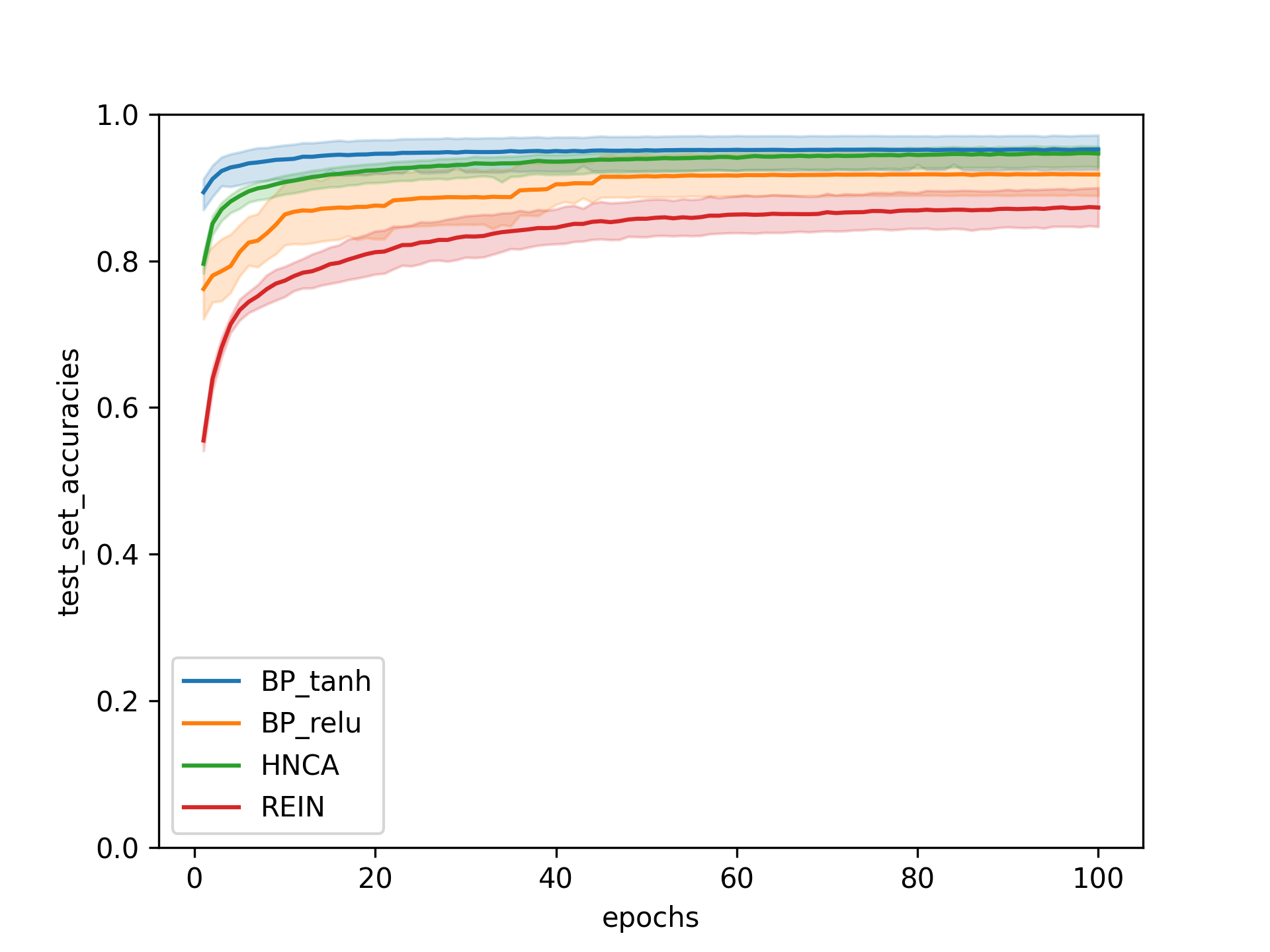}
\caption{Learning curves for best learning-rate for 1 hidden layer.}
\end{subfigure}
\hfill
\begin{subfigure}{0.3\textwidth}
\includegraphics[width=\textwidth]{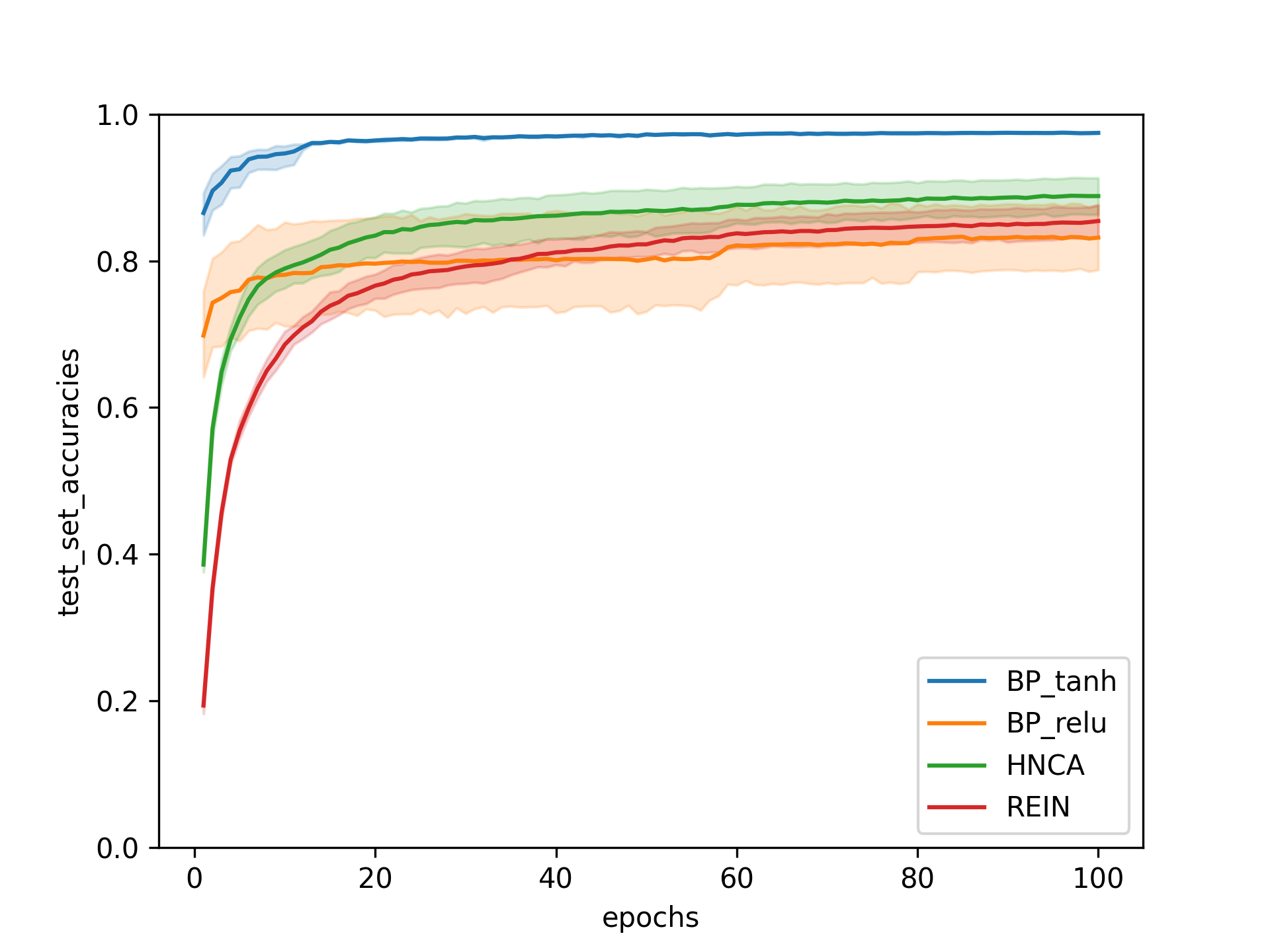}
\caption{Learning curves for best learning-rate for 2 hidden layers.}
\end{subfigure}
\hfill
\begin{subfigure}{0.3\textwidth}
\includegraphics[width=\textwidth]{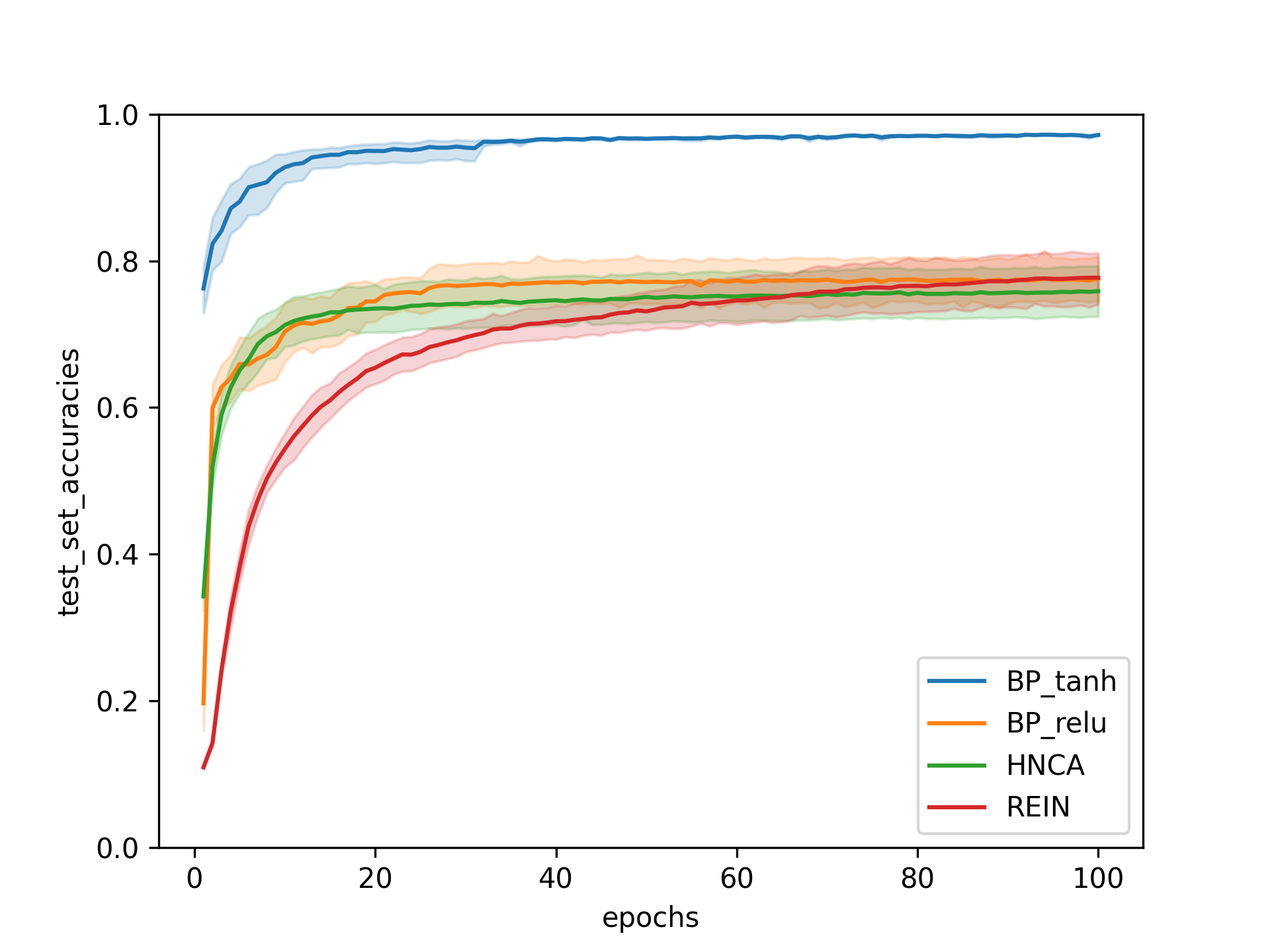}
\caption{Learning curves for best learning-rate for 3 hidden layers.}
\end{subfigure}
\caption{Learning curves and learning-rate sensitivity for HNCA and baselines on contextual bandit MNIST. Green curves are HNCA, red are REINFORCE, blue are backprop with sigmoid activations, and orange are backprop with ReLU activations. The architecture is a small neural network with 64 units per layer with different numbers of hidden layers. All plots show 10 random seeds with error bars showing 95\% confidence intervals. In order to show the fastest convergence among settings where final performance is similar, the best learning-rate is taken to be the highest learning-rate that is no more than one standard error from the learning-rate that gives the highest final accuracy.}\label{results}
\end{figure*}
\textbf{Experiments.}
We evaluate HNCA on MNIST~\citep{lecun2010mnist}, formulated as a contextual bandit. At each training step, the model outputs a class and receives a reward of $1$ only if the class is correct and $0$ otherwise. 

Our architecture consists of a stochastic feedforward neural network with 1,2 or 3 hidden layers, each with 64 Bernoulli nodes, followed by a softmax output layer. For training, we use batched versions of the algorithms in Appendix~\ref{implementation_details} with batchsize 16. We map the output of the Bernoulli nodes to one or negative one, instead of one or zero, as we found this significantly improved performance. For completeness, results using a zero-one mapping can be found in Appendix~\ref{exp_zero-one}.

As a baseline, we use the REINFORCE estimator with the same architecture. As two other baselines, we use backpropagation with an analogous deterministic architecture with tanh and ReLU activations trained using REINFORCE with the policy gradient of the network taken as a whole. All models, both deterministic and stochastic, use Glorot initialization~\citep{glorot2010understanding}. All models are trained for a total of 100 epochs with a batchsize of 16. For each architecture we sweep a wide range of learning-rates in powers of 2.

The results are shown in Figure~\ref{results}. As expected, we find that HNCA generally improves on REINFORCE, with the reduced variance facilitating stability at larger learning-rates, and leading to significantly faster learning. We also found that HNCA was generally competitive with the determinisitic network with ReLU activations. However, with tanh activations the deterministic network drastically outperforms all other tested approaches in this domain. This is a rather surprising result given the conventional wisdom that ReLU activation is superior. We speculate that this may be due to tanh allowing more flexible exploration, offering a benefit in the contextual bandit setting relative to ReLU where saturation may lead to premature convergence. However, given that this is not the main focus of this work we do not investigate this further. It would likely be possible to improve the performance of each approach through regularization, increased layer width, and other enhancements. Here, we choose to demonstrate the relative benefit of HNCA over REINFORCE in a simple setting.

\textbf{Conclusion.}
We introduce HNCA, a novel technique for computing unbiased gradient estimates for stochastic neural networks. HNCA works by estimating the gradient for each neuron based on how it contributes to the output of it's immediate children. We prove that HNCA provides lower variance gradient estimates compared to REINFORCE, and provide a computationally efficient pseudo-code implementation with complexity on the same order as backpropagation. We also experimentally demonstrate the benefit of HNCA over REINFORCE on a contextual bandit version of MNIST. 

Note that, unlike backpropagation, HNCA does not propagate credit information multiple steps. It only leverages information on how a particular node impacts its immediate children to produce a variance reduced gradient estimate. Extending this to propagate information multiple steps, while maintaining computational efficiency, appears nontrivial as a node's impact on descendants multiple steps away depends on intermediate nodes in a complex way. Nonetheless, there may be ways to extend this approach to propogate limited information muliple steps and we see this as a very interesting direction for future work.
\newpage
\printbibliography
\newpage

\appendix
\section{The Local REINFORCE Estimator is Unbiased}\label{local_REIN_unbiased}
Here we show that the local REINFORCE estimator $\hat{G}^{RE}_{\Phi}=\frac{\partial\log(\pi_\Phi(\Phi|B(\Phi)))}{\partial\theta_\Phi}R$ is an unbiased estimator of the gradient of the expected reward with respect to $\theta_\Phi$.
\begin{align*}
     \E[\hat{G}^{RE}_{\Phi}]&=\E\left[\frac{\partial\log(\pi_\Phi(\Phi|B(\Phi)))}{\partial\theta_\Phi}R\right]\\
     &=\sum_{b}\P(B(\Phi)=b)\E\left[\frac{\partial\log(\pi_\Phi(\Phi|B(\Phi)))}{\partial\theta_\Phi}R\middle|B(\phi)=b\right]\\
     &=\sum_{b}\P(B(\Phi)=b)\sum_\phi\pi_\Phi(\phi|b)\frac{\partial\log(\pi_\Phi(\phi|b))}{\partial\theta_\Phi}\E\left[R\middle|B(\Phi)=b,\Phi=\phi\right]\\
     &=\sum_{b}\P(B(\Phi)=b)\sum_\phi\frac{\partial\pi_\Phi(\phi|b)}{\partial\theta_\Phi}\E\left[R\middle|B(\Phi)=b,\Phi=\phi\right]\\
     &\stackrel{(a)}{=}\frac{\partial}{\partial\theta_\Phi}\sum_{b}\P(B(\Phi)=b)\sum_\phi\pi_\Phi(\phi|b)\E\left[R\middle|B(\Phi)=b,\Phi=\phi\right]\\
     &=\frac{\partial\E[R]}{\partial\theta_\Phi},
\end{align*}

where $(a)$ follows from the fact that the probability of the parents of $\Phi$, $\P(B(\Phi)=b))$, does not depend on the parameters $\Theta$ controlling $\Phi$ itself, nor does the expected reward once conditioned on $\Phi$. 

\section{Any Gradient Estimator Based on an Action-Value Estimator Obeying Equation~\ref{action_value_estimator} is Unbiased}\label{action_value_estiamtor_unbiased}
Assume we have access to an action value $\hat{Q}_\Phi(\phi)$ obeying Equation~\ref{action_value_estimator} and construct a gradient estimator $\hat{G}_{\Phi}=\sum_{\phi}\frac{\partial\pi_\Phi(\phi|B(\Phi))}{\partial\theta_\Phi}\hat{Q}_\Phi(\phi)$ as specified by Equation~\ref{PG_estimator}, then we can rewrite $\E[\hat{G}_{\Phi}]$ as follows:
\begin{align*}
     \E[\hat{G}_{\Phi}]&=\E\left[\sum_{\phi}\frac{\partial\pi_\Phi(\phi|B(\Phi))}{\partial\theta_\Phi}\hat{Q}_\Phi(\phi)\right]\\
     &=\sum_{b}\P(B(\Phi)=b)\sum_{\phi}\frac{\partial\pi_\Phi(\phi|b)}{\partial\theta_\Phi}\E\left[\hat{Q}_\Phi(\phi)\middle|B(\phi)=b\right]\\
     &\stackrel{(a)}{=}\sum_{b}\P(B(\Phi)=b)\sum_{\phi}\frac{\partial\pi_\Phi(\phi|b)}{\partial\theta_\Phi}\E[R|B(\Phi)=b,\Phi=\phi]\\
     &=\frac{\partial}{\partial\theta_\Phi}\sum_{b}\P(B(\Phi)=b)\sum_\phi\pi_\Phi(\phi|b)\E\left[R\middle|B(\Phi)=b,\Phi=\phi\right]\\
     &=\frac{\partial\E[R]}{\partial\theta_\Phi}.
\end{align*}
where $(a)$ follows from the assumption that $\hat{Q}_\Phi(\phi)$ obeys Equation~\ref{action_value_estimator}.
\newpage
\section{Examples where Assumption~\ref{parent_of_child} Holds and is Violated}\label{parent_of_child_examples}
\begin{figure*}[!h]
\centering
\begin{subfigure}[t]{0.3\textwidth}
\includegraphics[width=\textwidth]{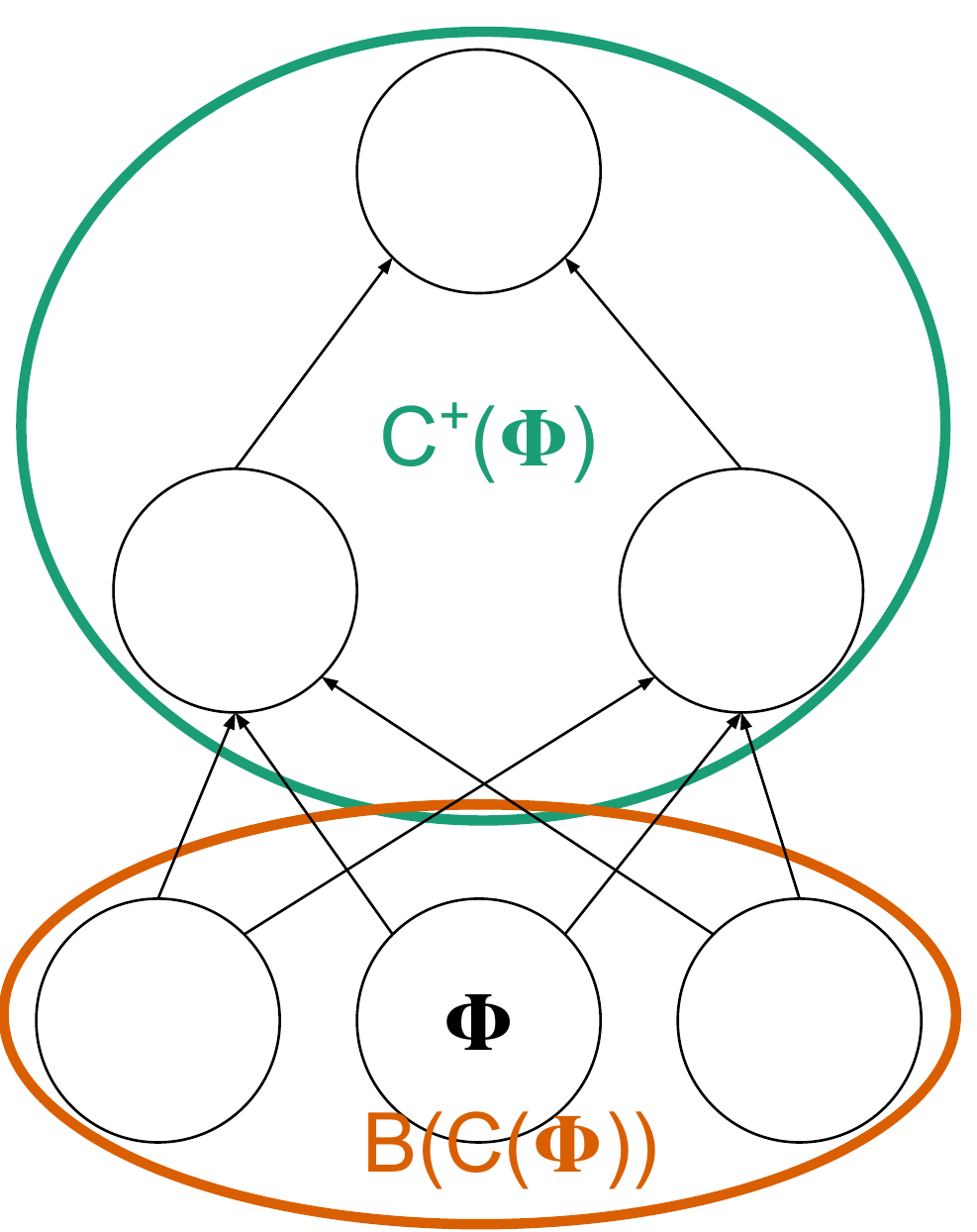}
\caption{An example where assumption~\ref{parent_of_child} holds.}
\label{obeyed}
\end{subfigure}
\hspace{0.13\textwidth}
\begin{subfigure}[t]{0.3\textwidth}
\includegraphics[width=\textwidth]{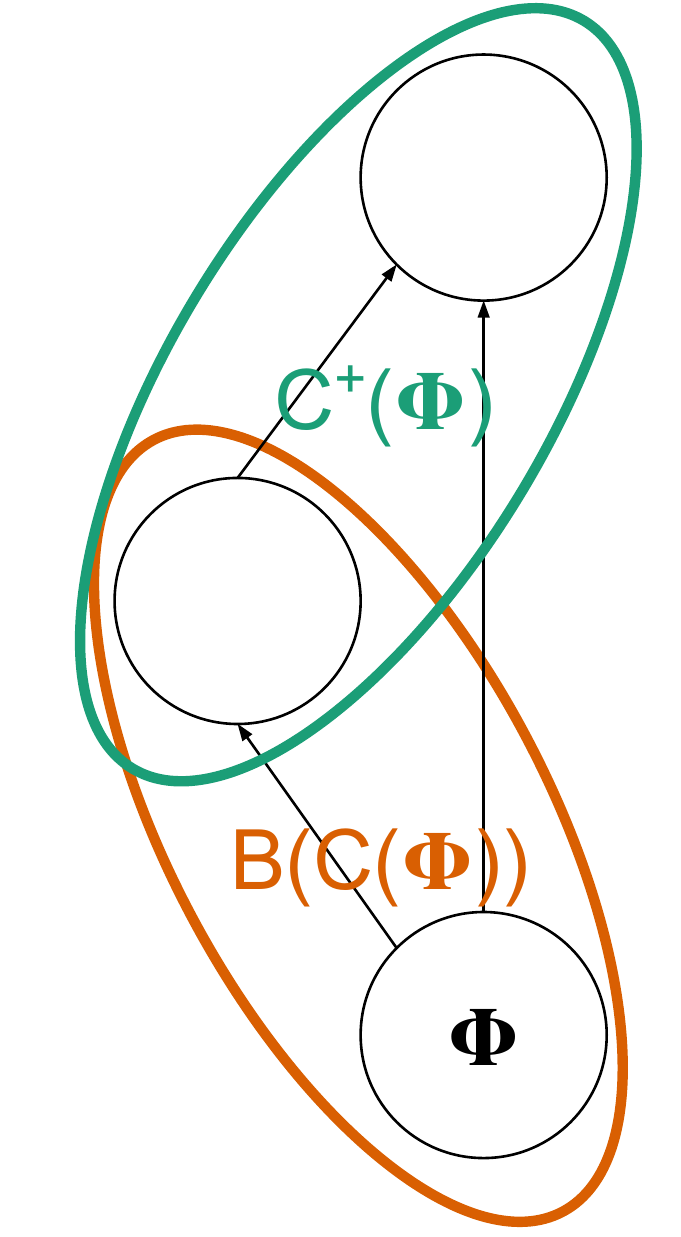}
\caption{An example where assumption~\ref{parent_of_child} fails to hold.}
\label{violated}
\end{subfigure}
\caption{}\label{obeyed_and_violated}
\end{figure*}

Figure~\ref{obeyed_and_violated} illustrates the meaning of assumption~\ref{parent_of_child} by showing a simple example where it holds, and another where it does not.

\section{The HNCA Action Value Estimator has Lower Variance than the Reinforce Estimator}\label{HNCA_action_value_low_var}
Here we provide the proof of Theorem~\ref{reduced_variance}.
\begingroup
\def\thetheorem{\ref{reduced_variance}}
\begin{theorem}
$\Var(\hat{Q}^{HNCA}_{\Phi}(\Phi)|B(\phi)=b)\leq \Var(\hat{Q}^{RE}_{\Phi}(\Phi)|B(\phi)=b).$
\end{theorem}
\addtocounter{theorem}{-1}
\endgroup
\begin{proof}
The proof follows from the law of total variance. First, note that by inverting up to step $(b)$ in Equation~\ref{HNCA_derivation} we know:
\begin{equation*}
     \hat{Q}^{HNCA}_\Phi(\phi)=\E\left[\frac{\ind(\Phi=\phi)}{\pi_\Phi(\phi|B(\Phi))}R\middle|C(\Phi),B(\Phi),B(C(\Phi))\setminus\Phi,R\right].
\end{equation*}
Now apply the law of total variance to rewrite the variance of the reinforce estimator as follows:
\begin{align*}
    \Var&(\hat{Q}^{RE}_{\Phi}(\Phi)|B(\phi)=b)\\
    &=\Var\left(\frac{\ind(\Phi=\phi)}{\pi_{\Phi}(\phi|B(\phi))}R\middle|B(\phi)=b\right)\\
    &=\begin{multlined}[t]
    \E\left[\Var\left(\frac{\ind(\Phi=\phi)}{\pi_{\Phi}(\phi|B(\phi))}R\middle|C(\Phi),B(\Phi),B(C(\Phi))\setminus\Phi,R\right)\middle|B(\phi)=b\right]\\
    +\Var\left(\E\left[\frac{\ind(\Phi=\phi)}{\pi_{\Phi}(\phi|B(\phi))}R\middle|C(\Phi),B(\Phi),B(C(\Phi))\setminus\Phi,R\right]\middle|B(\phi)=b\right)
    \end{multlined}\\
    &\geq \Var\left(\E\left[\frac{\ind(\Phi=\phi)}{\pi_{\Phi}(\phi|B(\phi))}R\middle|C(\Phi),B(\Phi),B(C(\Phi))\setminus\Phi,R\right]\middle|B(\phi)=b\right)\\
    &=\Var(\hat{Q}^{HNCA}(\Phi)|B(\phi)=b).
\end{align*}
\end{proof}

\section{Variance of the HNCA Gradient Estimator}\label{HNCA_gradient_low_var}
Theorem~\ref{reduced_variance} shows that the HNCA action value estimator $\hat{Q}^{HNCA}_\Phi(\Phi)$ has lower variance than $\hat{Q}^{RE}_\Phi(\Phi)$. In this section we discuss how this impacts the variance of the associated gradient estimator $\hat{G}_{\Phi}=\sum_{\phi}\frac{\partial\pi_\Phi(\phi|B(\Phi))}{\partial\theta_\Phi}\hat{Q}_\Phi(\phi)$. We can write this using the law of total variance as follows:
\begin{equation*}
    \Var(\hat{G}_{\Phi})=\E\left[\Var\left(\hat{G}_{\Phi}\middle|B(\Phi)\right)\right]+\Var\left(\E\left[\hat{G}_{\Phi}\middle|B(\Phi)\right]\right).
\end{equation*}
$\E\left[\hat{G}_{\Phi}\middle|B(\Phi)\right]$ is the same for both estimators so we will focus on $\E\left[\Var\left(\hat{G}_{\Phi}\middle|B(\Phi)\right)\right]$. Let $\Cov$ represent covariance.
\begin{equation}\label{expected_variance}
    \E\left[\Var\left(\hat{G}_{\Phi}\middle|B(\Phi)\right)\right]=\begin{multlined}[t]\sum_{b}\P(B(\Phi)=b)\Biggl(\sum_{\phi}\left(\frac{\partial\pi_\Phi(\phi|B(\Phi))}{\partial\theta_\Phi}\right)^2\Var(\hat{Q}_\Phi(\phi)|B(\Phi))\\
    +\sum_{\phi}\sum_{\phi^{\prime}\neq\phi}\left(\frac{\partial\pi_\Phi(\phi|B(\Phi))}{\partial\theta_\Phi}\frac{\partial\pi_\Phi(\phi^\prime|B(\Phi))}{\partial\theta_\Phi}\right)\Cov(\hat{Q}_\Phi(\phi),\hat{Q}_\Phi(\phi^\prime)|B(\Phi))\Biggr).
    \end{multlined}
\end{equation}
We have already shown that $\Var(\hat{Q}_\Phi(\phi)|B(\Phi))$ is lower for $\hat{Q}^{HNCA}_\Phi(\phi)$. Let us now look at $\Cov(\hat{Q}_\Phi(\phi),\hat{Q}_\Phi(\phi^\prime)|B(\Phi))$. For $\hat{Q}^{RE}_\Phi(\phi)$ only one $\phi$ takes nonzero value at a time, hence the covariance can be expressed as:
\begin{align*}
    \Cov(\hat{Q}^{RE}_\Phi(\phi),\hat{Q}^{RE}_\Phi(\phi^\prime)|B(\Phi))&=-\E[\hat{Q}^{RE}_\Phi(\phi)|B(\Phi)]\E[\hat{Q}^{RE}_\Phi(\phi^\prime)|B(\Phi)]\\
    &=-\E[R|B(\Phi),\Phi=\phi]\E[R|B(\Phi),\Phi=\phi^\prime].
\end{align*}
For $\hat{Q}^{HNCA}_\Phi(\phi)$ we can express the covariance as follows:
\begin{align*}
    \Cov&(\hat{Q}^{HNCA}_\Phi(\phi),\hat{Q}^{HNCA}_\Phi(\phi^\prime)|B(\Phi))=\begin{multlined}[t]\E[\hat{Q}^{HNCA}_\Phi(\phi)\hat{Q}^{HNCA}_\Phi(\phi^\prime)|B(\Phi)]\\
    -\E[\hat{Q}^{HNCA}_\Phi(\phi)|B(\Phi)]\E[\hat{Q}^{HNCA}_\Phi(\phi^\prime)|B(\Phi)]
    \end{multlined}\\
    &=\begin{multlined}[t]\E\left[\frac{\P\left(C(\Phi)\middle|B(C(\Phi))\setminus\Phi,\Phi=\phi\right)\P\left(C(\Phi)\middle|B(C(\Phi))\setminus\Phi,\Phi=\phi^\prime\right)}{\P(C(\Phi)|B(C(\Phi))\setminus\Phi,B(\Phi))^2}R^2|B(\Phi),\Phi=\phi\right]\\
    -\E\left[R|B(\Phi),\Phi=\phi]\E[R|B(\Phi),\Phi=\phi^\prime\right].
    \end{multlined}
\end{align*}
Note that the first term in this expression is always positive while the second is equal to the covariance expression for $\hat{Q}^{RE}_\Phi(\phi)$. Thus, $ \Cov(\hat{Q}^{HNCA}_\Phi(\phi),\hat{Q}^{HNCA}_\Phi(\phi^\prime)|B(\Phi))\geq\Cov(\hat{Q}^{RE}_\Phi(\phi),\hat{Q}^{RE}_\Phi(\phi^\prime)|B(\Phi))$ for all $\phi,\phi^\prime$.  Putting this all together and looking at Equation~\ref{expected_variance} we can conclude that $\Var(\hat{G}^{HNCA}_{\Phi})\leq\Var(\hat{G}^{RE}_{\Phi})$ as long as $\frac{\partial\pi_\Phi(\phi|B(\Phi))}{\partial\theta_\Phi}\frac{\partial\pi_\Phi(\phi^\prime|B(\Phi))}{\partial\theta_\Phi}$ is always negative. This is the case for Bernouli neurons, since changing any given parameter in this case will increase the probability of one actions and decrease the probability of the other. Here we use Bernoulli neurons in all but the final layer, where we cannot apply $\hat{G}^{HNCA}_{\Phi}$ anyways since $C(\hat\Phi)=\emptyset$. For more complex parameterizations (including softmax) this will not hold exactly. However, speaking roughly, since the policy gradients still need to sum to one over all $\phi$, the gradients with respect to different actions will tend to be negatively correlated, and thus $\hat{G}^{HNCA}_{\Phi}$ will tend to be lower variance.

\section{Details on Computationally Efficient Implementation}\label{implementation_details}
\begin{figure}[h]
\begin{minipage}{0.46\textwidth}
\begin{algorithm}[H]
Receive $\vec{x}$ from parents\\
$l=\vec{\theta}\cdot\vec{x}+b$\\
$p=\sigma(l)$\\
$\phi\sim \textit{Bernoulli}(p)$\\
Pass $\phi$ to children\\
Receive $\vec{q}_1,\vec{q}_0,R$ from children\\
$q_1=\prod_{i}\vec{q}_1[i]$\\
$q_0=\prod_{i}\vec{q}_0[i]$\\
$\bar{q}=pq_1+(1-p)q_o$\\
$\vec{l}_1=l+\vec{\theta}\odot(1-\vec{x})$\\
$\vec{l}_0=l-\vec{\theta}\odot\vec{x}$\\
$\vec{p}_1=\sigma(\vec{l}_1)$\\
$\vec{p}_0=\sigma(\vec{l}_0)$\\
Pass $\vec{p}_1,\vec{p}_0,R$ to parents\\
$\vec{\theta}=\vec{\theta}+\alpha\sigma^{\prime}(l)\vec{x}\left(\frac{q_1-q_0}{\bar{q}}\right)R$\\
$b=b+\alpha\sigma^{\prime}(l)\left(\frac{q_1-q_0}{\bar{q}}\right)R$
\caption{HNCA algorithm for Bernoulli hidden neuron}\label{bernoulli_alg}
\end{algorithm}
\end{minipage}
\hfill
\begin{minipage}{0.46\textwidth}
\begin{algorithm}[H]
Receive $\vec{x}$ from parents\\
$\vec{l}=\Theta\vec{x}+\vec{b}$\\
$\vec{p}=\frac{\exp{\vec{l}}}{\sum_i\exp{\vec{l}[i]}}$\\
Output $\phi\sim \vec{p}$\\
Receive $R$ from environment\\
\For{all $i$}{
$L_1[i]=\vec{l}+\Theta[i]\odot(1-\vec{x})$\\
$L_0[i]=\vec{l}-\Theta[i]\odot\vec{x}$
}
$\vec{p}_1=\frac{\exp{L_1[\phi]}}{\sum_i\exp{L_1[i]}}$\\
$\vec{p}_0=\frac{\exp{L_0[\phi]}}{\sum_i\exp{L_0[i]}}$\\
Pass $\vec{p}_1,\vec{p}_0,R$ to parents\\
\For{all $i$}{
$\Theta[i]=\Theta[i]+\alpha\vec{x}(\ind(\phi=i)-\vec{p}[i])R$\\
$b[i]=b[i]+\alpha(\ind(\phi=i)-\vec{p}[i])R$
}
\caption{HNCA algorithm for Softmax output neuron}\label{softmax_alg}
\end{algorithm}
\end{minipage}
\end{figure}
The computational complexity of HNCA depends on how difficult it is to compute $\P\left(C(\Phi)\middle|B(C(\Phi))\setminus\Phi,\Phi=\phi\right)$ and $\P(C(\Phi)|B(C(\Phi))\setminus\Phi,B(\Phi))$ in Equation~\ref{HNCA_estimator}. When the individual neurons use a sufficiently simple parameterization, the method can be implemented as a backpropagation-like message passing procedure. In this case, the overall computation is proportional to the number of connections, as in backpropagation itself.

In our experiments, we will apply HNCA to solve a classification task formulated as a contextual bandit (the agent must guess the class from the input and receives a reward of 1 only if the guess is correct, otherwise it does not receive the true class).

Our stochastic neural network model will consist of a number of hidden layers, wherein each neuron outputs according to a Bernoulli distribution. The policy of each Bernoulli neuron is parametrized as a linear function of it's inputs, followed by a sigmoid activation. The policy of the output layer is a distribution over class labels, parameterized as a softmax. We now separately highlight the implementations for the softmax output and Bernoulli hidden nodes.

Algorithm~\ref{bernoulli_alg} shows the implementation of HNCA for the Bernoulli hidden nodes. The pseudo-code provided is for Bernoulli nodes with a zero-one mapping, but is straightforward to modify to use negative one and one instead, as we do in our main experiments. Lines 1-5 implement the forward pass. The forward pass simply takes input from the parents, uses it to compute the fire probability $p$ and samples $\phi\in\{0,1\}$. The backward pass receives two vectors of probabilities $\vec{q}_1$ and $\vec{q}_0$, each with one element for each child of the node. A given element represents $\vec{q}_{0/1}[i]=\P\left(C_i\middle|B(C_i)\setminus\Phi,\Phi=0/1\right)$ for a given child $C_i\in C(\Phi)$. Lines 7 and 8 take the product of all these child probabilities to compute $\P\left(C(\Phi)\middle|B(C)\setminus\Phi,\Phi=0/1\right)$. Note that computing $\P\left(C(\Phi)\middle|B(C)\setminus\Phi,\Phi=0/1\right)$ in this way is made possible by Assumption~\ref{parent_of_child}. Due to Assumption~\ref{parent_of_child}, no $C_i\in C(\Phi)$ can influence another $C_j\in C(\Phi)$ via a downstream path. Thus $\P\left(C(\Phi)\middle|B(C)\setminus\Phi,\Phi=0/1\right)=\prod_i\P\left(C_i\middle|B(C_i)\setminus\Phi,\Phi=0/1\right)$. Line 9 uses $\vec{q}_{0/1}[i]$ along with the fire probability to compute $\bar{q}=\P(C(\Phi)|B(C(\Phi))\setminus\Phi,B(\Phi))$.

Line 10-13 use the already computed logit $l$ to efficiently compute a vector of probabilities $\vec{p}_1$ and $\vec{p}_0$ where each element corresponds to a counterfactual fire probability if all else was the same but a given parent's value was fixed to 1 or 0. Here, $\odot$ represents the element-wise product. Note, that computing this in this way only requires compute time on the order of the number of parents (while naively computing each counterfactual probability from scratch would require time on the order of the number of children squared). Line 14 passes the nessesary information to the node's children. Lines 15 and 16 finally update the parameter using $\hat{G}^{HNCA}_{\Phi}$ with learning-rate hyperparameter $\alpha$.

Algorithm~\ref{softmax_alg} shows the implementation for the softmax output node. Note that the output node itself uses the REINFORCE estimator in its update, as it has no children. Nonetheless, the output node still needs to provide information to its parents, which use HNCA. Lines 1-4 implement the forward pass, in this case producing an integer $\phi$ drawn from the possible classes. Lines 6-11 compute counterfactual probabilities of the given output class conditional on fixing the value of each parent. Note that $\Theta[i]$ refers to the $i_{th}$ row of the matrix $\Theta$. In this case, computing these counterfactual probabilities requires computation on the order of the number of parents, times the number of possible classes. Line 12 passes the necessary information back to the parents. Lines 13-16 update the parameters according to $\hat{G}^{RE}_{\Phi}$.

Overall, the output node requires computation proportional to the number of parents times the number of classes, this is the same the number of parameters in that node. Similarly, the hidden nodes each require computation proportional to it's number of parents, which again is the same as the number of parameters. Thus the overall computation required for HNCA is on the order of the number of parameters, the same order as doing a forward pass, and the same order as backpropagation. 

It's worth noting that, in principle, HNCA is also more parallelizable than backpropagation. Since no information from a node's children is needed to compute the information passed to its parents, the backward pass could be done in parallel across all layers. However, this can also be seen as a limitation of the proposed approach since the ability to condition on information further upstream could lead to further variance reduction.

\section{Experiments with Zero-One Output Mapping}~\label{exp_zero-one}
\begin{figure*}[h]
\centering
\begin{subfigure}{0.3\textwidth}
\includegraphics[width=\textwidth]{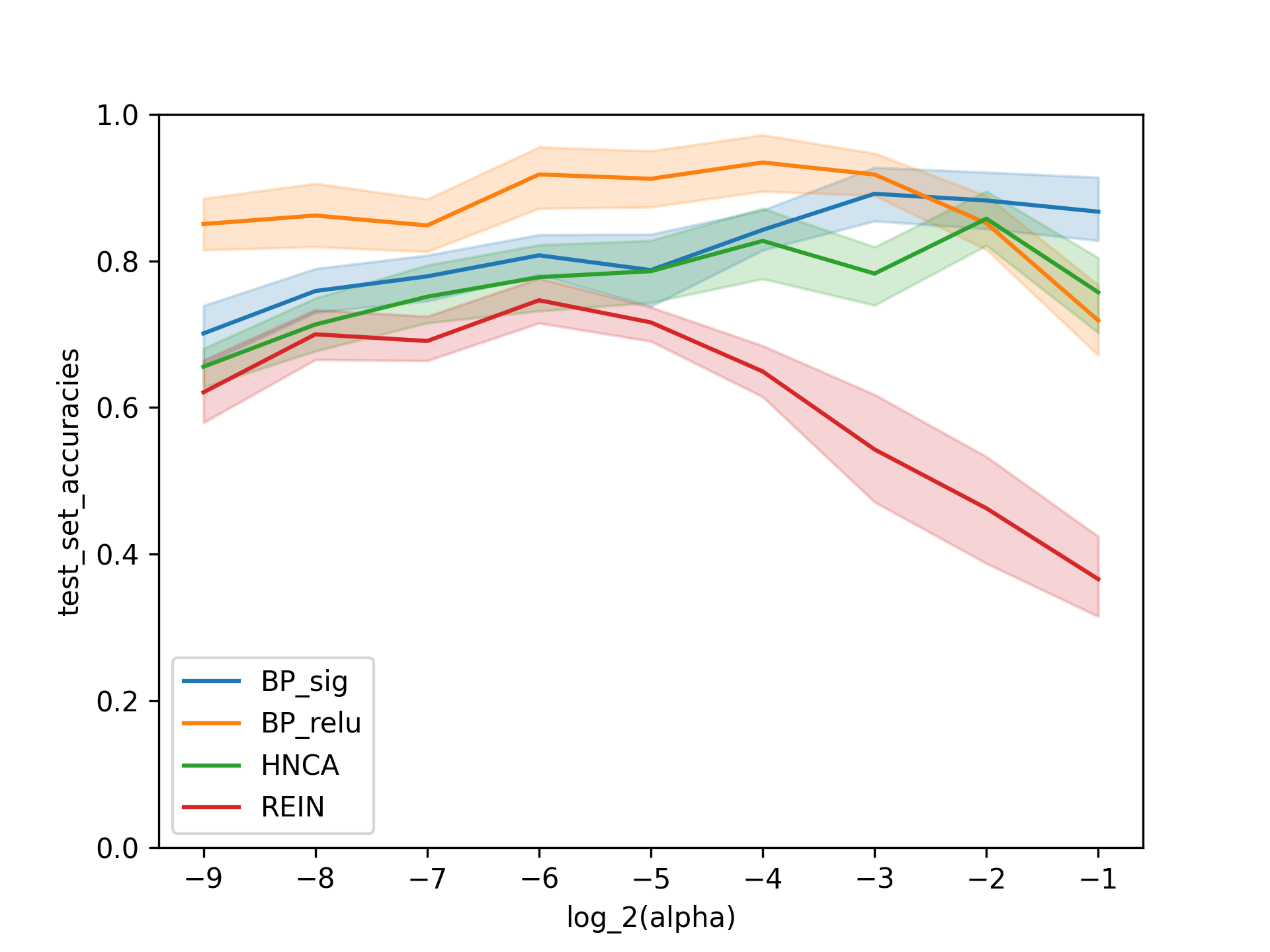}
\caption{Learning-rate sensitivity curves for 1 hidden layer.}
\label{SCE_oscillatory}
\end{subfigure}
\hfill
\begin{subfigure}{0.3\textwidth}
\includegraphics[width=\textwidth]{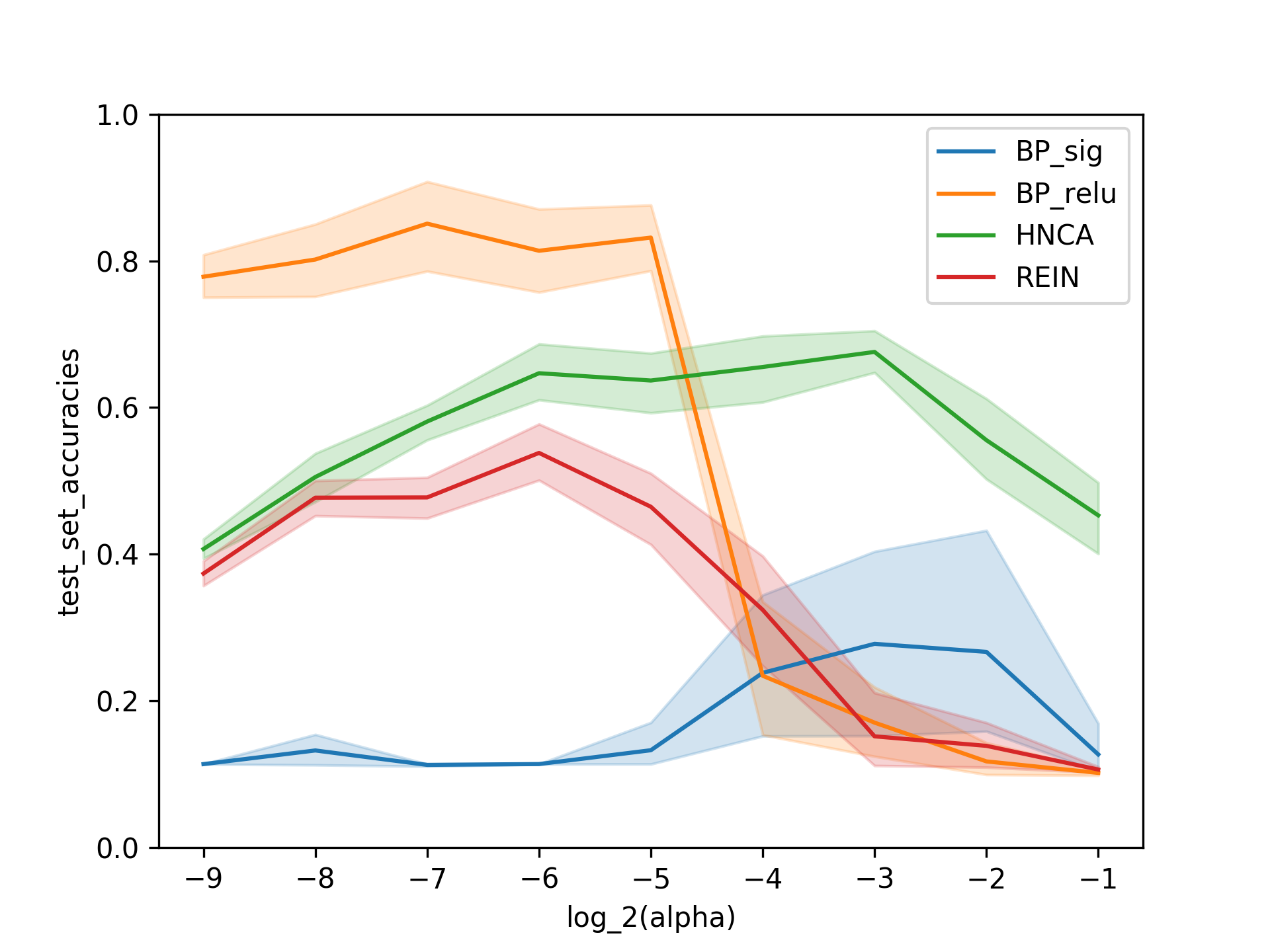}
\caption{Learning-rate sensitivity curves for 2 hidden layers.}
\label{SCE_oscillatory}
\end{subfigure}
\hfill
\begin{subfigure}{0.3\textwidth}
\includegraphics[width=\textwidth]{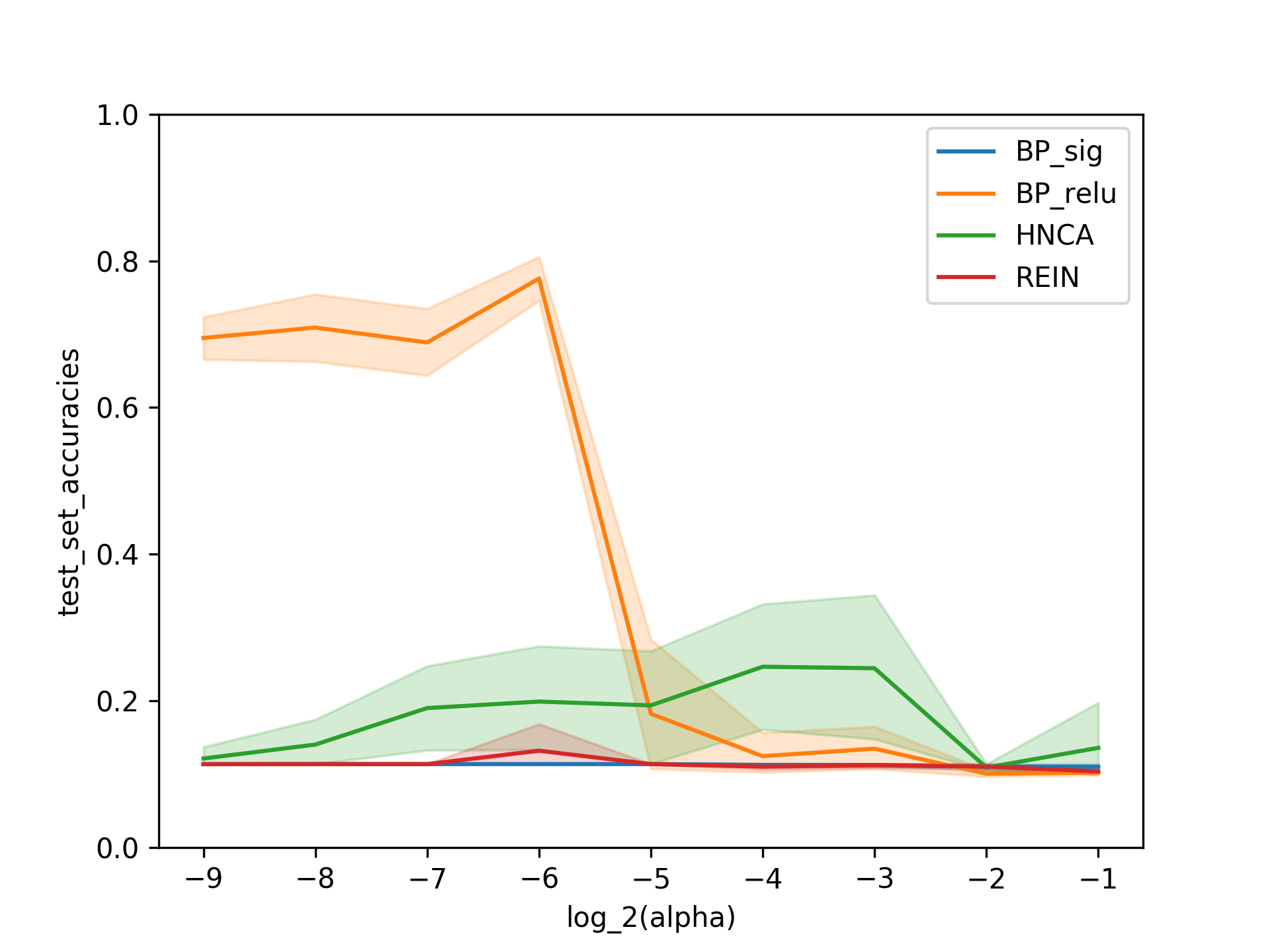}
\caption{Learning-rate sensitivity curves for 3 hidden layers.}
\end{subfigure}
\linebreak
\begin{subfigure}{0.3\textwidth}
\includegraphics[width=\textwidth]{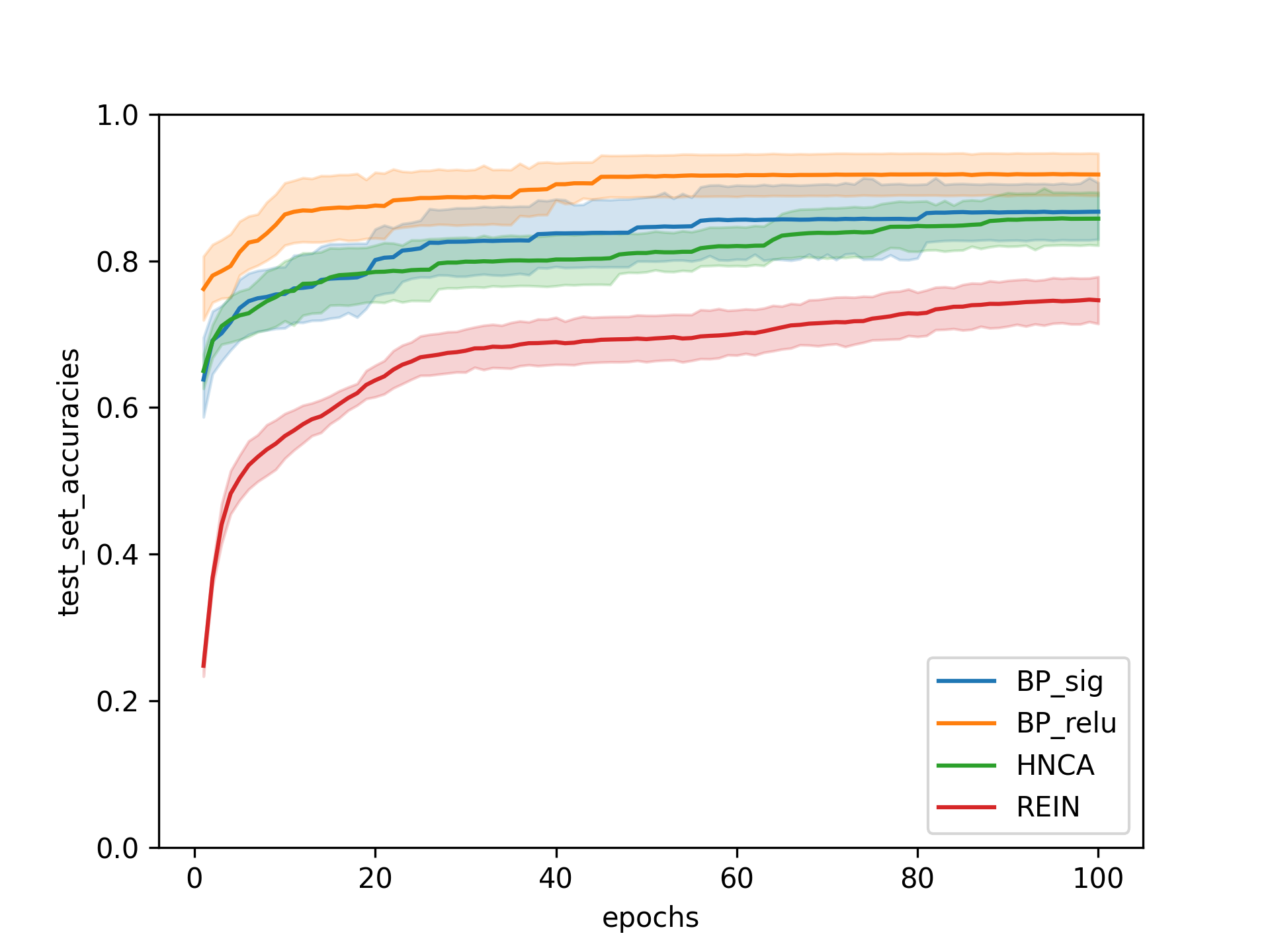}
\caption{Learning curves for best learning-rate for 1 hidden layer.}
\end{subfigure}
\hfill
\begin{subfigure}{0.3\textwidth}
\includegraphics[width=\textwidth]{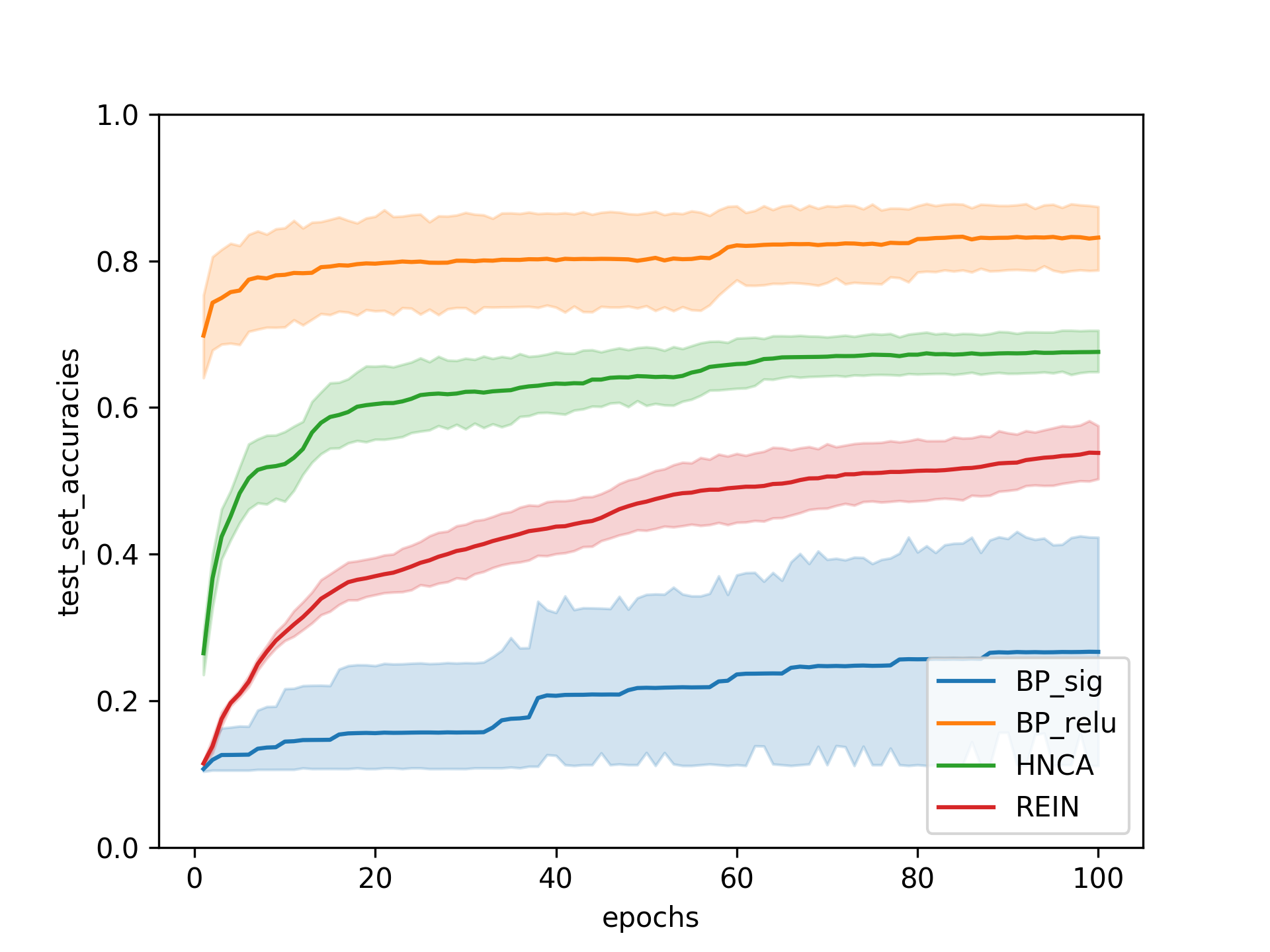}
\caption{Learning curves for best learning-rate for 2 hidden layers.}
\end{subfigure}
\hfill
\begin{subfigure}{0.3\textwidth}
\includegraphics[width=\textwidth]{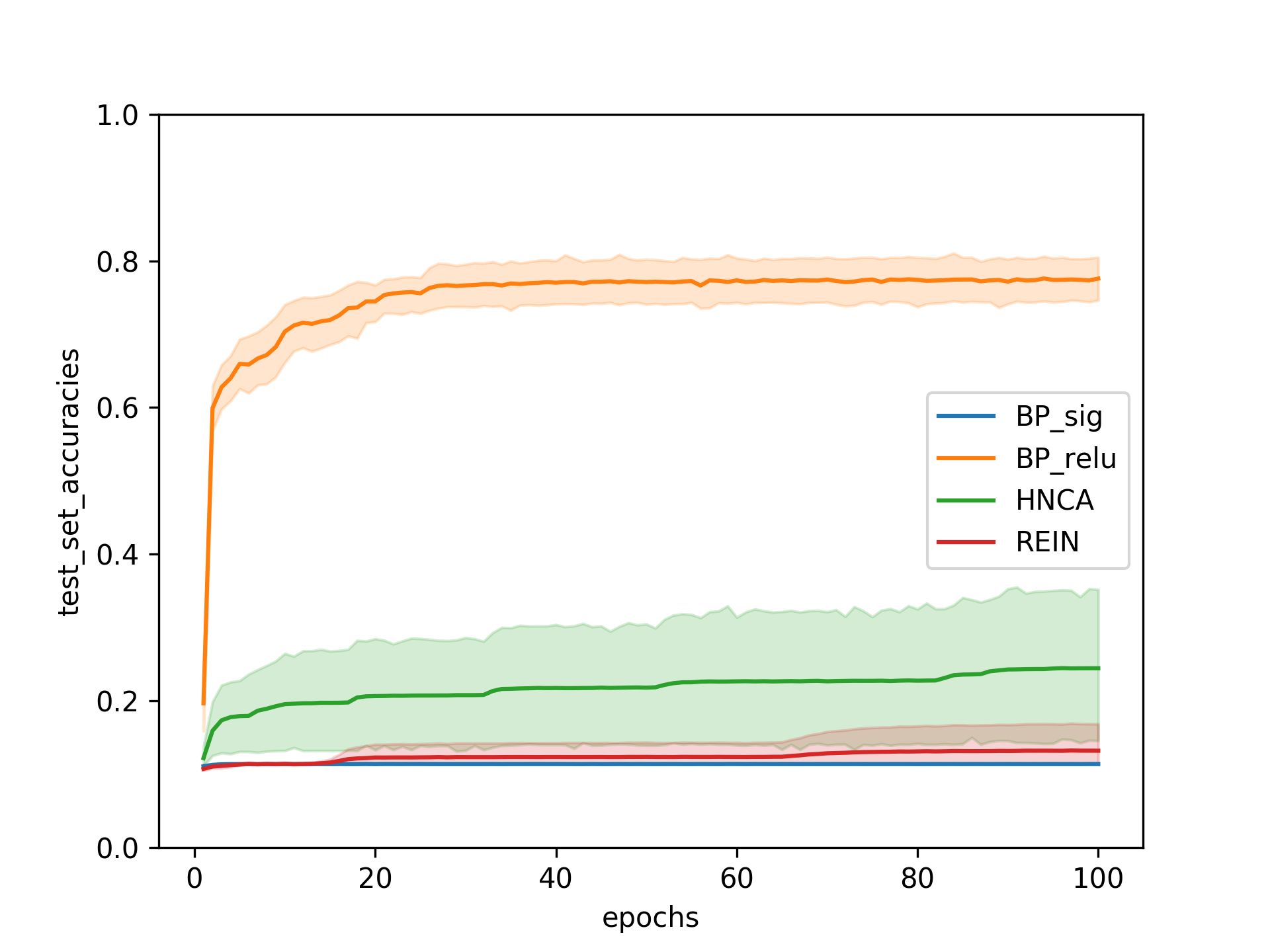}
\caption{Learning curves for best learning-rate for 3 hidden layers.}
\end{subfigure}
\caption{Learning curves and learning-rate sensitivity for HNCA (with a zero-one mapping) and baselines on contextual bandit MNIST. Green curves are HNCA, red are REINFORCE (each with a zero-one mapping), blue are backprop with sigmoid activations, and orange are backprop with ReLU activations. The architecture is a small neural network with 64 units per layer with different numbers of hidden layers. All plots show 10 random seeds with error bars showing 95\% confidence intervals. In order to show the fastest convergence among settings where final performance is similar, the best learning-rate is taken to be the highest learning-rate that is no more than one standard error from the learning-rate that gives the highest final accuracy.}\label{results_zero-one}
\end{figure*}

In the main body of the paper our experiments mapping the binary output of Bernoulli neurons to one or negative one. We found that this worked much better than mapping to zero or one. In Figure~\ref{results_zero-one}, we show the results with of REINFORCE and HNCA with a zero-one mapping. We also use a sigmoid activation instead of tanh backpropagation baseline, as this offers a closer analog to the stochastic zero-one mapping. In each case, the use of a zero-one mapping significantly hurt performance. Replacing tanh with sigmoid or the backpropagation baseline makes the difference between significantly outperforming ReLU and barely learning in the 2 and 3 hidden layers cases.

\end{document}